\documentclass[conference]{IEEEtran}
\IEEEoverridecommandlockouts
\usepackage{cite}
\usepackage{amsmath,amssymb,amsfonts}
\usepackage{algorithmic}
\usepackage{graphicx}
\usepackage{textcomp}
\usepackage{xcolor}
\usepackage{multirow}
\usepackage{colortbl}
\usepackage[utf8]{inputenc}
\usepackage[russian,english]{babel}
\usepackage{pgfplots}
\usepackage{hyperref}
\usepackage{subcaption}
\usepackage{hhline}
\DeclareUnicodeCharacter{2212}{−}
\usepgfplotslibrary{groupplots,dateplot}
\usetikzlibrary{patterns,shapes.arrows}
\pgfplotsset{compat=newest}

\definecolor{rankcolor1}{rgb}{0.42274509803921567,0.684075355632449,0.8398923490965013}
\definecolor{avgrankcolor1}{rgb}{0.41636293733179547,0.3190003844675125,0.639923106497501}
\definecolor{avgrankcolor2}{rgb}{0.6214532871972318,0.606074586697424,0.7855440215301807}
\definecolor{avgrankcolor3}{rgb}{0.8568396770472895,0.8566551326412918,0.9224913494809688}

\newcommand{\Mapping}{\mathcal{M}}
\newcommand{\Training}{\mathcal{M}_{T}}
\newcommand{\KG}{\mathcal{KG}}
\newcommand{\Ent}{\mathcal{E}}
\newcommand{\AttRel}{\mathcal{A}}
\newcommand{\EntRel}{\mathcal{R}}
\newcommand{\Lit}{\mathcal{L}}
\newcommand{\Triples}{\mathcal{T}}
\newcommand{\AttTrip}{\mathcal{T}_{A}}
\newcommand{\RelTrip}{\mathcal{T}_{R}}
\newcommand{\FeatureVector}{\mathcal{V}}
\newcommand{\FVAttr}{\mathcal{V}_{A}}
\newcommand{\FVEmb}{\mathcal{V}_{E}}

\newcommand{\ApproachBoth}{\textsc{EAGER}_{A\mathbin\Vert E}}
\newcommand{\ApproachAttr}{\textsc{EAGER}_{A}}
\newcommand{\ApproachEmb}{\textsc{EAGER}_{E}}

\begin{document}

\title{EAGER: Embedding-Assisted Entity Resolution for Knowledge Graphs
\thanks{This work was supported by the German Federal Ministry of Education and Research (BMBF, 01/S18026A-F) by funding the competence center for Big Data and AI "ScaDS.AI Dresden/Leipzig". Some computations have been done with resources of Leipzig University Computing Center.
}
}

\author{
\IEEEauthorblockN{Daniel Obraczka}\IEEEauthorblockA{Leipzig University\\obraczka@informatik.uni-leipzig.de}
\and
\IEEEauthorblockN{Jonathan Schuchart}\IEEEauthorblockA{Leipzig University\\schuchart@informatik.uni-leipzig.de}
\and
\IEEEauthorblockN{Erhard Rahm}\IEEEauthorblockA{Leipzig University\\rahm@informatik.uni-leipzig.de}
}

\maketitle

\begin{abstract}
Entity Resolution (ER) is a constitutional part for integrating different knowledge graphs in order to identify  entities  referring to the same real-world object. A promising approach is the use of graph embeddings for ER in order to determine the similarity of entities based on the similarity of their graph neighborhood.  The similarity computations for such embeddings translates to calculating the distance between them in the embedding space which is comparatively simple. 
However, previous work has shown that the use of graph embeddings alone is not sufficient to achieve high ER quality. We therefore propose a more comprehensive ER approach for knowledge graphs called EAGER (\textit{E}mbedding-\textit{A}ssisted Knowledge \textit{G}raph \textit{E}ntity \textit{R}esolution) to flexibly utilize both the similarity of graph embeddings and attribute values within a supervised machine learning approach. 
We evaluate our approach on 23 benchmark datasets with differently sized and structured knowledge graphs and use hypothesis tests to ensure statistical significance of our results.
Furthermore we compare our approach with state-of-the-art ER solutions, where our approach yields competitive results for table-oriented ER problems and shallow knowledge graphs but much better results for deeper knowledge graphs.  
\end{abstract}


\section{Introduction}
Knowledge Graphs (KGs) store real-world facts in machine-readable form. This is done by making statements about entities in triple form $(entity, property, value)$. For example the triple \texttt{(Get\_Out, director, Jordan\_Peele)} tells us that the director of the movie "Get Out" is "Jordan Peele". Such structured information can be used for a variety of tasks such as recommender systems, question answering and semantic search. For many KG usage forms including question answering it is beneficial to integrate KGs from different sources. An integral part of this integration is entity resolution (ER), where the goal is to find entities which refer to the same real-world object.

\begin{figure}[]
    \centering
    \includegraphics[width=\columnwidth]{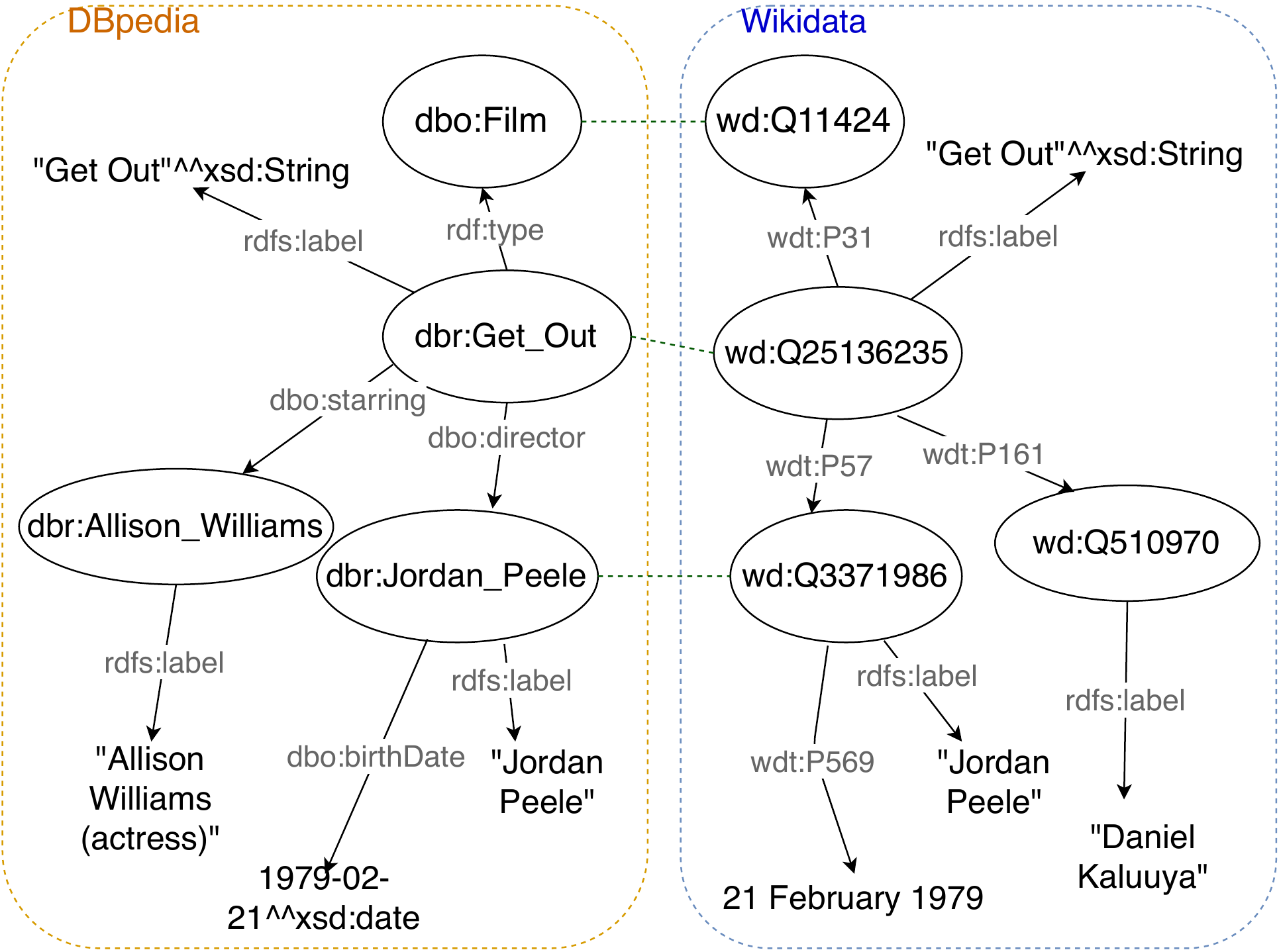}
    \caption{Subgraphs of DBpedia and Wikidata. Green dashed lines show entities that should be matched. Some URIs are shortened for brevity.}
    \label{fig:matching_example}
\end{figure}

Existing ER systems mostly focus on matching entities of one specific entity type (e.g. publication, movie, customer etc.) and assume matched schemata for this entity type.
This proves challenging when trying to use these systems for ER in KGs typically consisting of many entity types with heterogeneous attribute (property) sets. This is illustrated by the example in Figure~\ref{fig:matching_example} showing two simple movie-related KG subgraphs to be matched with each other.
We observe there are  entities of different types (film, director, actor) and different attributes with heterogeneous value representations (e.g., birth date values "1979-02-21" in DBpedia and "21 Febuary 1979" in Wikidata for two matching director entities). Moreover, we see that matching entities such as the movie "Get Out" have different URIs and  differently named edges referring to properties and related entities, e.g. \texttt{rdf:type} vs. \texttt{wdt:P31}. These aspects make a traditional schema (property) matching as a means to simplify ER very challenging so that entity resolution for KGs should ideally not depend on it. Given that URIs and property names may not show any similarity it becomes apparent that the graph structure and related entities should be utilized in the ER process, e.g., to consider the movie label and  director to match movies.   

A promising way to achieve this in a generic manner, applicable to virtually any entity type, is the use of graph embeddings. 
 By encoding the entities of the KGs into a low-dimensional space such approaches alleviate the obstacles posed by the aforementioned KG heterogeneities. Capturing the topological and semantic relatedness of entities in a geometric embedding space enables the use of these embeddings as inputs for machine learning (ML) algorithms.
The performance of graph embedding approaches for ER has been recently studied by Sun et. al~\cite{Sun2020}. However, as they point out, most approaches focus on refining the embedding process, while ER mostly consists of finding the nearest neighbors in the embedding space. Hence, the use of graph embeddings has to be tailored to the ER task for good effectiveness. We build on the findings of ~\cite{Sun2020} and investigate 
 the usefulness of learned graph embeddings as input for ML classifiers for entity resolution. While there are different settings for KG  integration, such as enhancing a given KG or KG fusion, we  focus here on the simple ER setting, i.e., finding matching entities in two data sources. The resulting match mappings can then be used for applications such as question answering or as input for KG fusion. 

In this paper, we propose and comprehensively evaluate  the first (to our knowledge)  \textit{graph embedding supported} ER system named  \textbf{EAGER}: \textbf{E}mbedding \textbf{A}ssisted Knowledge \textbf{G}raph \textbf{E}ntity \textbf{R}esolution. 
It uses both knowledge graph embeddings and attribute similarities as inputs for an ML classifier for entity resolution. EAGER utilizes different kinds of graph embeddings, specifically the ones that performed best in \cite{Sun2020}, as well as different ML classifiers. We comprehensively evaluate the match effectiveness and runtime efficiency of EAGER with 
using graph embeddings and attribute similarities either alone or in combination for 23 datasets of varying size and structure. We also compare 
the different graph embeddings and classifiers with each other to identify good default configurations. 
We further provide a comparison of EAGER with state-of the-art ER approaches, namely Magellan~\cite{Magellan} and DeepMatcher~\cite{DeepMatcher}. All our results are analyzed using hypothesis tests to ensure statistical significance of our findings.   

We begin by presenting related work followed by a description of EAGER. In Section~\ref{sec:datasets} the used datasets are presented including a new benchmark dataset from the movie domain. Our evaluation is presented in Section~\ref{sec:eval} and we end with conclusions and future work in Section~\ref{sec:conclusion}.

\section{Related Work and Background}
Entity resolution has attracted a significant amount of research, sometimes under different names such as record linkage~\cite{Domingos2004MultiRelationalRL,Elfeky2002TAILORAR}, link discovery~\cite{Volz2009SilkA,Sherif2017WombatA} or deduplication~\cite{Sarawagi2002InteractiveDU}.
In the following we can only present some relevant ER approaches. We refer the reader to surveys and books like ~\cite{Elmagarmid2007DuplicateRDSurvey,Nentwig2017Survey, Christen2012DataMatching} for a more thorough overview. 
Traditional ER approaches rely on learning distance- or similarity-based measures and then use a threshold or classifier to decide about whether two entities are the same.
These classifiers can be unsupervised~\cite{Ngomo2013UnsupervisedLO,Nikolov2012UnsupervisedLO}, supervised~\cite{Sherif2017WombatA,Isele2012LearningEL} or employ active learning~\cite{Sarawagi2002InteractiveDU,Ngomo2012EAGLEEA}. For example the Magellan Framework~\cite{Magellan} provides supervised ml classifiers and provides extensive guides for the entire ER process.
Recently, deep learning has seen some success in certain settings. DeepER~\cite{DeepER} and DeepMatcher~\cite{DeepMatcher} provide a variety of different architectures and among other aspects, such as attribute similarities, use word embeddings as inputs for these networks. Both frameworks have shown that especially for unstructured textual data deep learning can outperform existing frameworks.

Collective ER approaches try to overcome the limitations of the more conventional attribute-based methods.
This paradigm uses the relationships between entities as additional information and in some cases even considers previous matching decisions in the neighborhood.
Bhattacharya and Getoor~\cite{Bhattacharya2006CollectiveER} show that using the neighborhood of potential match candidates in addition to attribute-based similarity is especially useful for data with many ambiguous entities.
SiGMa~\cite{LacosteJulien2013SIGMaSG} uses an iterative graph propagation algorithm relying on relationship information as well as attribute-based similarity between graphs to integrate large-scale knowledge bases.
Pershina et al.~\cite{Pershina2015HolisticEM} propagate similarities using Personalized PageRank and are able to align industrial size knowledge graphs.
Zhu et al.~\cite{Zhu2016UnsupervisedER} reformulate entity resolution as multi-type graph summarization problem and use attribute-based similarity as well as structural similarity, i.e. connectivity patterns in the graph.

More recently the use of graph embeddings has been shown promising for the integration of KGs. An overview of currently relevant approaches that solely rely on embedding techniques can be found in~\cite{Sun2020}, some of these techniques have been used in this work and will be discussed in more detail in Section \ref{ApproachEmbeddings}.

Knowledge graph embedding (KGE) models typically aim to capture the relationship structure of each entity in latent vector representations in order to be used for further downstream applications. For a good overview of current knowledge graph embedding approaches we refer the reader to a recent survey from Ali et al. \cite{Ali2020PyKeen}.
A widely used basic technique are translational models, such as TransE by Bordes et al. \cite{Bordes2013TransE} and its various proposed improvements TransH \cite{Wang2014TransH}, TransR \cite{Lin2015TransR} and TransD \cite{Ji2015TransD}. Translational models interpret a relationship as a translation from its head entity to its tail entity. Note that translational models also embed relationship names and would therefore benefit from consistent vocabularies (schemata) across knowledge graphs.
Trouillon et al. \cite{trouillon2016complex} used complex valued vectors in order to better capture anti-symmetric relationships, similar to an idea proposed by \cite{sun2019rotate} which restricts these complex representations to the unit circle.
Based on the influential Graph Convolutional Network (GCN) model by Kipf and Welling \cite{kipf2016GCN} for ordinary graphs, Schichtkrull et al. \cite{Schichtkrull2018RGCN} used relationship specific weight matrices to capture relations as well as the neighborhood structure of each entity.

EAGER aims to combine the two generally separate ER approaches of entity embedding based techniques and traditional attribute based methods in KGs. We show that our approach is viable for both real world KGs and artificial shallow KGs that are based on tabular data as EAGER does not rely on additional schema matching or any structural assumptions about the entities.

\section{An overview of EAGER}
\begin{figure*}[]
    \centering
    \includegraphics[width=\linewidth]{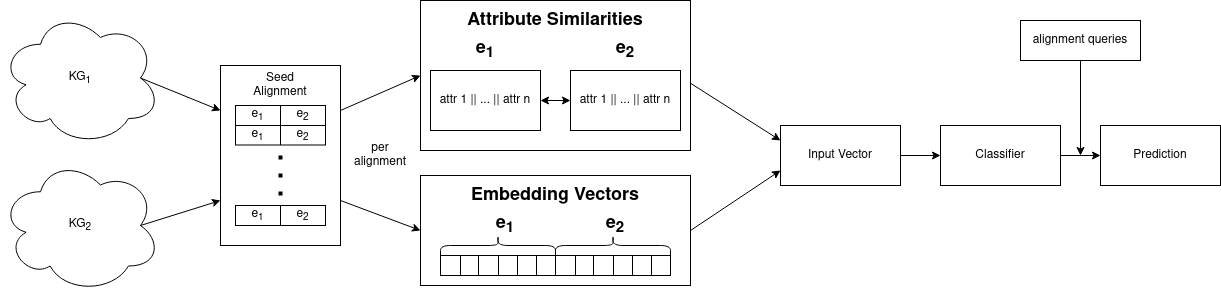}
    \caption{Schematic summary of EAGER}
    \label{fig:approach}
\end{figure*}
In this section we present an overview  of the EAGER approach for ER in knowledge graphs and the specific approaches and configurations we will evaluate. 
We start with a formal definition of the ER problem and an overview of the EAGER workflow. We then discuss problems when calculating attribute similarities in heterogeneous KGs and our use of graph embeddings. We finish the section by describing the different variants of EAGER we intend to evaluate and present the machine learning classifiers that we use. We close this chapter by discussing the prediction step.
\subsection{Problem statement}
As  stated in the introduction, KGs are constructed by triples in the form of $(entity, property, value)$, where $property$ can be either a attribute property or a relationship and $value$ a literal or another entity, respectively.  
Therefore, a KG is a tuple $\KG = (\Ent, \EntRel, \AttRel, \Lit, \Triples)$, where $\Ent$ is the set of entities, $\AttRel$ the set of attribute properties, $\EntRel$ the set of relationship properties, $\Lit$ the set of literals and $\Triples$ is the set of triples. 
We distinguish attribute triples $\AttTrip$ and relationship triples $\RelTrip$, where $\AttTrip : \Ent \times \AttRel \times \Lit$ are triples connecting entities and literals, e.g. \texttt{(dbr:Jordan\_Peele, dbo:birthDate, "1979-02-21")} and $\RelTrip : \Ent \times \EntRel \times \Ent$ connect entities, e.g. \texttt{(dbr:Get\_Out, dbo:director, dbr:Jordan\_Peele)} as seen in Figure~\ref{fig:matching_example}.
Our goal is to find a mapping between entities of two KGs. More formally, we aim to find $\Mapping = \{(e_{1},e_{2}) \in \Ent_1 \times \Ent_2 | e_1 \equiv e_2 \}$, where $\equiv$ refers to the equivalence relation. 
Furthermore, we assume we are provided with a subset of the mapped entities $\Training \subseteq \Mapping$ as training data, which is also sometimes referred to as seed alignment in the literature.
\subsection{Overview}
The remaining chapter is dedicated to illustrate how our approach tackles entity resolution in heterogeneous KGs.
A schematical overview can be found in Figure~\ref{fig:approach}.
Given two KGs $\KG_1,\KG_2$ and a set of initial matches $\Training$ we create a feature vector $\FeatureVector$ for each match $(e_1, e_2) \in \Training$ to train a machine learning classifier. Additionally to the positive matches provided in $\Training$ we sample negative examples by sampling random pairs $(e_1, e_2) \notin \Training$ to create a balanced set of positive and negative examples.
After the training step the classifier then acts as an oracle to answer specific alignment queries, i.e. entity pairs, in order to make a prediction.
In the following we present our approach in more detail.
\subsection{Attribute Similarities}
Since schemata across different KGs may differ wildly, creating a schema matching before ER in heterogeneous KGs is difficult and can introduce additional sources for error. While matching attributes by hand is possible for datasets with a low number of attributes this is not possible for large KGs, where more sophisticated approaches are necessary. Keeping the focus on the matching process, we chose to concatenate all attribute values of each entity into a single string and used 3 similarity measures for comparisons: Levenshtein, Generalized Jaccard with an Alphanumeric Tokenizer, which returns the longest strings of alphanumeric characters, and Trigrams with the Dice coefficient.
This results in three separate features that can be used as input to a classifier. Note, that EAGER is generally not bound to any specific distance/similarity measures and any other features that can be derived from two strings can be used.
\subsection{Graph Embeddings}\label{ApproachEmbeddings}
Given that the focus of this study lies not on the creation of embeddings itself, our approach can take any entity embeddings that are embedded in the same space. Since most KG embedding frameworks are not specialized for ER, we use OpenEA\footnote{https://github.com/nju-websoft/OpenEA} which was developed by Sun et al. for their 2020 benchmark study\cite{Sun2020}. It offers a variety of embedding approaches and embeds entities into the same space. Specifically, we chose three of the best approaches of said study, namely BootEA, MultiKE and RDGCN: 
\subsubsection{BootEA}
Sun et al. in 2018 \cite{Sun2018BootEA} based their approach on the TransE model and combined it with elaborate bootstrapping and negative sampling techniques to improve performance. TransE aims to find an embedding function $\phi$ that minimizes $||\phi(e_h) + \phi(r) - \phi(e_t)||$ for any $(e_h, r, e_t) \in \RelTrip$. Bootstrapping is done by additionally sampling likely matching entities (resampled every few epochs based on the current model) in order to increase the effective seed alignment size. Additionally, negative relationship tuples are sampled and resampled every few epochs based on the current model in order to improve the distinction between otherwise similar entities. Since TransE is an unsupervised model, Sun et al. proposed a new objective function which incorporates both the original objective function of TransE and the likelihood of two entities from different KGs matching. Thus making use of the seed alignment.
\subsubsection{MultiKE}
In order to also incorporate more than just relational information, Zhang et al. \cite{Zhang2019MultiKE} proposed a flexible model which combines different views on each entity. Here, the name attribute, relations and all remaining attributes are embedded separately, using pre-trained word2vec word embeddings \cite{mikolov2013word2vec} for names and a variation on TransE for relations. Attribute embeddings are obtained by training a convolutional neural network taking the attribute and attribute value as input. All three embedding vectors are then combined into a single unified embedding space. In this approach the two knowledge graphs are treated as one combined graph where entities from the seed alignment are treated as equal.
\subsubsection{RDGCN}
Different to the aforementioned approaches, Wu et al. \cite{Wu2019RDGCN} proposed a new technique using two constructed conventional graphs and the GCN model by Kipf and Welling with highways. Instead of learning embeddings for entities and relations within one graph, RDGCN constructs a primary entity graph and a dual relationship graph in order to alternate the optimization process between the two. That way, the relationship representations from the dual graph are used to optimize the entity representations from the primal graph and vice versa by applying a graph attention mechanism. As the actual neigborhood information of each entity is not fully exploited in this case, Wu et al. showed that feeding the resulting entity representations into a GCN can help significantly improve the overall embedding quality.
\subsection{Combinations}
As the aim of our study is to investigate, whether combining entity embeddings with attribute similarities is superior to using either on their own, we present three different variants of our approach, that only differ in the construction of their feature vector $\FeatureVector$: 
\begin{itemize}
    \item  $\ApproachBoth$, where $\FeatureVector = concat(\FVAttr, \FVEmb)$
    \item $\ApproachEmb$, where $\FeatureVector = \FVEmb$
    \item $\ApproachAttr$, where $\FeatureVector = \FVAttr$
\end{itemize}
Where $\FVAttr$ contains the attribute similarities, and $\FVEmb$ the embeddings. The $concat$ operation simply appends one vector to the other.
\subsection{Prediction}
The trained classifier is presented with alignment queries, i.e. pairs of entities that it will have to classify as match or non-match. Choosing these pairs is a non-trivial question since exploring all possible pairs would lead to a quadratic number of alignment queries relative to the KG size, which is not scalable to large datasets. Traditionally, blocking strategies are used to reduce the number of pairs by a linear factor. Due to the heterogeneous nature of KGs new strategies for this problem have to be found. An alternative could be to use the embeddings to find a number of nearest neighbors, which is a scalable solution since the triangle inequality in metric spaces can be exploited to reduce the number of comparisons for the neighborhood search.
Finding a good solution for this problem is however out of scope for our study and in the experiments we therefore use the test data to create prediction pairs, sampling negative examples randomly as done in the training step. More on our experimental setup can be found in Section~\ref{sec:evalsetup}.

\section{Datasets}
\label{sec:datasets}
To evaluate our approach we use multiple datasets that can generally be put into two categories: \textit{rich} and \textit{shallow} graph datasets. While the former are sampled from popular knowledge graphs and therefore contain a rich graph structure, i.e. lots of different relationships, the latter are derived from tabular data and have a very limited number of relationships.

\begin{table*}[]
\caption{Shallow graph datasets statistics}
\centering
\resizebox{\columnwidth}{!}{%
\begin{tabular}{|cc|cccccc|}
 \hline
Datasets & KGs & $|\EntRel|$ & $|\AttRel|$ & $|\RelTrip|$ & $|\AttTrip|$ & $|\Ent|$ & $|\Mapping|$\\
 \hline
\multirow{2}{*}{abt-buy} & abt & 3 & 4 & 2753 & 2998 & 1920 & \multirow{2}{*}{1097}\\
& buy & 4 & 4 & 4654 & 3480 & 2392 &\\ \hline
\multirow{2}{*}{amazon-google} & amazon & 4 & 4 & 8528 & 5802 & 4443 & \multirow{2}{*}{1300}\\
& google & 4 & 4 & 16429 & 12971 & 9749 &\\ \hline
\multirow{2}{*}{acm-dblp} & acm & 4 & 3 & 15007 & 5874 & 9190 & \multirow{2}{*}{2224}\\
& dblp & 4 & 3 & 16444 & 6041 & 10462 &\\ \hline
\multirow{2}{*}{dblp-gs} & dblp & 4 & 3 & 16017 & 5832 & 10256 & \multirow{2}{*}{5347}\\
& gs & 4 & 3 & 390579 & 190336 & 228211 &\\ \hline
\multirow{2}{*}{imdb-tmdb} & imdb & 3 & 13 & 17532 & 25723 & 5129 & \multirow{2}{*}{1978}\\
 & tmdb & 4 & 493 & 27903 & 24695 & 6056 & \\ \hline
\multirow{2}{*}{imdb-tvdb} & imdb & 3 & 13 & 17532 & 25723 & 5129 & \multirow{2}{*}{2488}\\
& tvdb & 3 & 350 & 15455 & 21430 & 7810 &\\ \hline
\multirow{2}{*}{tmdb-tvdb} & tmdb & 4 & 493 & 27903 & 24695 & 6056 & \multirow{2}{*}{2483}\\
 & tvdb & 3 & 350 & 15455 & 21430 & 7810 &\\
 \hline
\end{tabular}
}
\label{table:shallowgraphstatistics}
\end{table*}
\begin{table*}[]
\caption{Rich graph datasets statistics, adapted from~\cite{Sun2020}}
\centering
\begin{tabular}{|c|cc|ccccc|ccccc|}
 \hline
\multicolumn{2}{|c}{\multirow{2}{*}{Datasets}} & \multirow{2}{*}{KGs} & \multicolumn{5}{c|}{V1} & \multicolumn{5}{c|}{V2} \\
\cline{4-13}
\multicolumn{2}{|c}{}&& $|\EntRel|$ & $|\AttRel|$ & $|\RelTrip|$ & $|\AttTrip|$ & $|\Mapping|$ &$|\EntRel|$ & $|\AttRel|$ & $|\RelTrip|$ & $|\AttTrip|$ & $|\Mapping|$ \\
 \hline
\multirow{8}{*}{\rotatebox[origin=c]{90}{15K}}&\multirow{2}{*}{D-W} & DB & 248 & 342 & 38,265 & 68,258 &\multirow{2}{*}{15,000} & 167 & 175 & 73,983 & 66,813 &\multirow{2}{*}{15,000}\\
&& WD & 169 & 649 & 42,746 & 138,246 && 121 & 457 & 83,365 & 175,686 &\\ \cline{2-13}
&\multirow{2}{*}{D-Y} & DB & 165 & 257 & 30,291 & 71,716 & \multirow{2}{*}{15,000} & 72 & 90 & 68,063 & 65,100 &\multirow{2}{*}{15,000}\\
&& YG & 28 & 35 & 26,638 & 132,114 & & 21 & 20 & 60,970 & 131,151 &\\ \cline{2-13}
&\multirow{2}{*}{EN-DE} & EN & 215 & 286 & 47,676 & 83,755 & \multirow{2}{*}{15,000}& 169 & 171 & 84,867 & 81,988 &\multirow{2}{*}{15,000}\\
& & DE & 131 & 194 & 50,419 & 156,150 & &96 & 116 & 92,632 & 186,335 & \\  \cline{2-13}
&\multirow{2}{*}{EN-FR} & EN & 267 & 308 & 47,334 & 73,121 & \multirow{2}{*}{15,000} & 193 & 189 & 96,318 & 66,899 & \multirow{2}{*}{15,000} \\
&& FR & 210 & 404 & 40,864 & 67,167 & &166 & 221 & 80,112 & 68,779 &\\
\hline
\multirow{8}{*}{\rotatebox[origin=c]{90}{100K}}&\multirow{2}{*}{D-W} & DB & 413 & 493 & 293,990 & 451,011 &\multirow{2}{*}{100,000}& 318 & 328 & 616,457 & 467,103  &\multirow{2}{*}{100,000}\\
&& WD &261 & 874 & 251,708 & 687,860 & &239 & 760 & 588,203 & 878,219 &\\  \cline{2-13}
&\multirow{2}{*}{D-Y} & DB & 287 & 379 & 294,188 & 523,062 & \multirow{2}{*}{100,000} & 230 & 277 & 576,547 & 547,026 &\multirow{2}{*}{100,000}\\
&& YG & 32 & 38 & 400,518 & 749,787 & & 31 & 36 & 865,265 & 855,161 &\\ \cline{2-13}

&\multirow{2}{*}{EN-DE} & EN & 381 & 451 & 335,359 & 552,750 & \multirow{2}{*}{100,000}& 323 & 326 & 622,588 & 560,247& \multirow{2}{*}{100,000} \\
& & DE & 196 & 252 & 336,240 & 716,615 & &170 & 189 & 629,395 & 793,710& \\ \cline{2-13}
& \multirow{2}{*}{EN-FR} & EN & 400 & 466 & 309,607 & 497,729 & \multirow{2}{*}{100,000} & 379 & 364 & 649,902 & 503,922 & \multirow{2}{*}{100,000}\\
& & FR & 300 & 519 & 258,285 & 426,672 && 287 & 468 & 561,391 & 431,379 &\\ 
\hline
\end{tabular}
\label{table:richgraphstatistics}
\end{table*}

In this section we present the datasets used for our evaluation, starting with the shallow graph datasets, followed by the rich graph datasets.
\subsection{Shallow Graph Datasets}
To investigate how the interplay of attribute similarities and graph embeddings fares in settings with less dense KGs we transformed classical ER benchmark datasets and created a new benchmark dataset with multiple entity types. The classical ER datasets are taken from ~\cite{Kopcke2010} and transformed into simple KGs. Due to repurposing of these ER tasks we only have the gold standard for one entity type: publication for the benchmarks from the publication domain, and product from the datasets associated with e-commerce. 
To address this shortcoming we created a new benchmark from the movie domain, where the gold standard was hand-labeled for the five entity types \texttt{Person, Movie, TvSeries, Episode, Company}.
The movie datasets were created from three sources containing information about movies and tv series: IMDB\footnote{\url{https://www.imdb.com/}}, TheMovieDB\footnote{\url{https://www.themoviedb.org/}} and TheTVDB\footnote{\url{https://www.thetvdb.com/}}. 
We provide more details about the datasets in Table ~\ref{table:shallowgraphstatistics}. It is important to note, that the repurposed classical ER benchmark datasets have a very low number of different attributes, while the movie datasets are more rich in this respect. Also note that the number of entities in the knowledge graphs is different to the published number of products or articles for abt-buy, amazon-google, dblp-acm and dblp-scholar as the knowledge graphs contain additional entity types such as places, events, authors, brands and prices. We make the movie datasets publicly available for future research at \url{https://github.com/ScaDS/MovieGraphBenchmark}.

\subsection{Rich Graph Datasets}
In the study by Sun et al.~\cite{Sun2020} the authors provided datasets from DBpedia (DB), Wikidata (WD) and Yago (YG) that were sampled with the intention of properly emulating the graph structure of real-world KGs. To investigate several aspects that are relevant in the ER process they provide two versions of each linking task where V1 has dataset samples that are less dense than V2. Additionally, there is a small and large integration task with each dataset consisting of 15K and 100K entities, with the gold standard for each task containing 15K and 100K matches respectively. It is worth mentioning, that two of the ER tasks have a cross-lingual character with samples from the English (EN), French (FR) and German (DE) versions of DBpedia. The datasets show a variety of entity types. For example the 100K version of D-W (V2) has 91 different values for relationship triples with the property \texttt{dbo:type} in the DBpedia KG. These entity types have a wide range from movies and persons to geographical locations and corporations. Due to the sampling done by Sun et al. the type information is missing for most entities, making the real variety of entity types much larger. More details about the datasets are provided in Table~\ref{table:richgraphstatistics}.

\section{Evaluation} \label{sec:eval}
We discuss our results on the presented datasets, starting with a description of the experiment setup, followed by the results on the shallow and rich graph datasets, with a focus on investigating whether the use of attribute similarities in combination with knowledge graph embeddings is beneficial for the respective setting. Furthermore we compare our approach with state-of-the-art frameworks and present runtimes of our approach.
\subsection{Setup} \label{sec:evalsetup}
For the evaluation we use a 5-fold cross validation with a 7-2-1 split in accordance with \cite{Sun2020}: For each dataset pair the set of reference entity matches is divided into 70\% testing, 20\% training and 10\% validation. For each split we sample negative examples to create an equal share of positive and negative examples. The entire process is repeated 5 times to create 5 different folds.

For the OpenEA datasets the graph embeddings were computed using the hyperparameters given by the study of~\cite{Sun2020}. For all other datasets, apart from dblp-scholar, the *-15K parameter sets were used. For dblp-scholar, the *-100K parameters were applied as the scholar dataset contains more than 100,000 entities.
For the classifiers, Random Forest Classifier was used with 500 estimators and a Multi Layer Perceptron (MLP) was used with two hidden layers of size 200 and 20. Furthermore, MLP was trained using the Adam~\cite{kingma2017adam} optimizer with $\alpha = 10^{-5}$.

\subsection{Shallow Graph Datasets}
The results for the shallow datasets are displayed in Table~\ref{tab:shallowtable}. 
We display the average rank of each combination of input variant, embedding approach and classifier which is a number between 1 and 14 (since there are 14 possible combinations), where 1 would mean this combination achieves the best result for each dataset.

As expected there is too little information in the shallow datasets to produce good results with the embeddings alone. We can also see that $\ApproachBoth$ and $\ApproachAttr$ perform similarly. However there is an apparent difference in performance between the movie datasets and the others. Out of the three embedding approaches only $\ApproachBoth$ with MultiKE performs better than $\ApproachAttr$ on the movie datasets.
For the classical ER benchmarks using only $\FVAttr$ as input for either RF or MLP gives the best results overall. While MultiKE performs the second-worst for $\ApproachEmb$ it gives the best results when used in $\ApproachBoth$. Averaged over all shallow test datasets, $\ApproachAttr$ performs best and the Random Forest (RF) classifier was the most effective classifier reaching the lowest average ranks.  

\begin{table*}[t]
    \caption{Averaged F-measure on test set of shallow graph datasets. The best value in a row is highlighted}
    \label{tab:shallowtable}
    \centering
\begin{tabular}{|c|cc|cc|cc|cc|cc|cc|cc|}
\hline
\multirow{3}{*}{Dataset} & \multicolumn{6}{c|}{$\ApproachBoth$} & \multicolumn{2}{c|}{\multirow{2}{*}{$\ApproachAttr$}} & \multicolumn{6}{c|}{$\ApproachEmb$} \\
\cline{2-7}
\cline{10-15}
 & \multicolumn{2}{c|}{BootEA} & \multicolumn{2}{c|}{MultiKE} & \multicolumn{2}{c|}{RDGCN} & \multicolumn{2}{c|}{} & \multicolumn{2}{c|}{BootEA} & \multicolumn{2}{c|}{MultiKE} & \multicolumn{2}{c|}{RDGCN} \\
 \cline{2-15}
 &             MLP &                         RF &                        MLP &                         RF &    MLP &                         RF &                        MLP &                                                                                 RF &     MLP &      RF &     MLP &     RF &     MLP &      RF \\
\hline
abt-buy       &           0.885 &                      0.952 &                      0.958 &                      0.952 &  0.925 &                      0.920 &  {\cellcolor{rankcolor1}}0.968 &                                                                              0.965 &   0.623 &   0.648 &   0.383 &  0.655 &   0.650 &   0.661 \\
amazon-google &           0.751 &                      0.798 &                      0.789 &                      0.760 &  0.784 &                      0.768 &                      0.808 &                                                          {\cellcolor{rankcolor1}}0.817 &   0.631 &   0.646 &   0.571 &  0.645 &   0.638 &   0.665 \\
dblp-acm      &           0.995 &  {\cellcolor{rankcolor1}}0.997 &  {\cellcolor{rankcolor1}}0.997 &  {\cellcolor{rankcolor1}}0.997 &  0.995 &  {\cellcolor{rankcolor1}}0.997 &  {\cellcolor{rankcolor1}}0.997 &                                                          {\cellcolor{rankcolor1}}0.997 &   0.579 &   0.614 &   0.617 &  0.688 &   0.559 &   0.598 \\
dblp-scholar  &           0.993 &                      0.997 &                      0.994 &                      0.997 &  0.995 &                      0.996 &                      0.997 &                                                          {\cellcolor{rankcolor1}}0.998 &   0.562 &   0.588 &   0.537 &  0.576 &   0.547 &   0.571 \\
imdb-tmdb     &           0.967 &                      0.977 &  {\cellcolor{rankcolor1}}0.988 &                      0.984 &  0.969 &                      0.975 &                      0.979 &                                                                              0.980 &   0.874 &   0.859 &   0.911 &  0.913 &   0.874 &   0.873 \\
imdb-tvdb     &           0.938 &                      0.960 &  {\cellcolor{rankcolor1}}0.973 &                      0.967 &  0.940 &                      0.953 &                      0.965 &                                                                              0.960 &   0.821 &   0.786 &   0.873 &  0.844 &   0.807 &   0.792 \\
tmdb-tvdb     &           0.973 &                      0.977 &  {\cellcolor{rankcolor1}}0.983 &                      0.981 &  0.966 &                      0.977 &                      0.980 &                                                                              0.978 &   0.874 &   0.844 &   0.871 &  0.877 &   0.857 &   0.831 \\
\hline
Avg Rank      &           7.786 &                      4.143 &                     {\cellcolor{avgrankcolor3}} 2.929 &                      3.429 &  6.643 &                      5.571 &                      {\cellcolor{avgrankcolor2}}2.786 &  {\cellcolor{avgrankcolor1}}\textcolor{white}{2.714} &  11.929 &  11.857 &  11.714 &  9.714 &  12.214 &  11.571 \\
\hline
\end{tabular}
\end{table*}
\begin{figure*}[]
    \centering
    \includegraphics[width=\textwidth]{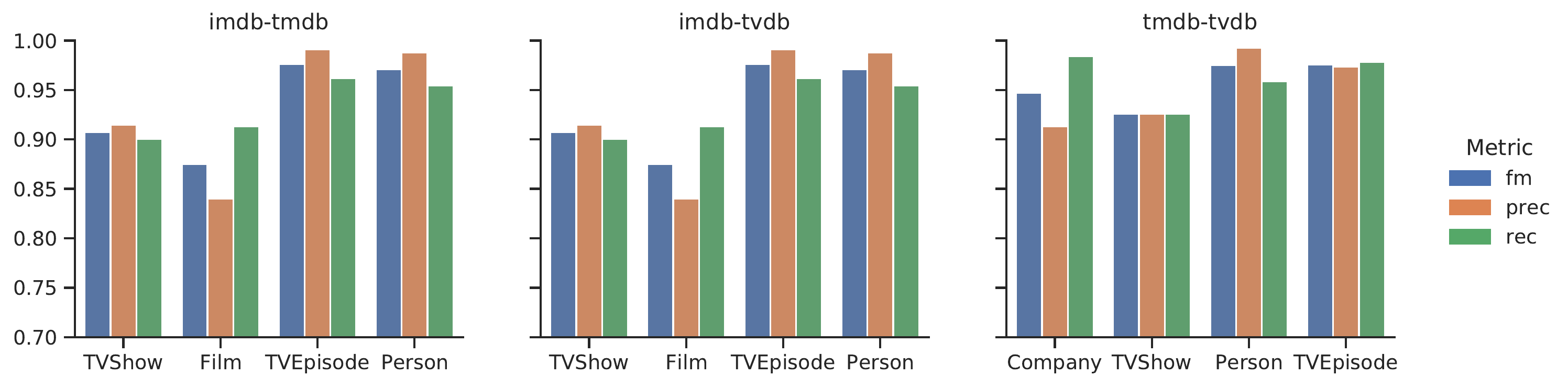}
    \caption{Averaged F-measure, Precision and Recall per Type on Movie Datasets using $\ApproachBoth$ with MLP}
    \label{fig:movietypes}
\end{figure*}

Looking at the movie datasets in more detail as shown in Figure~\ref{fig:movietypes}, we can see that there is a difference in performance depending on the entity type. In most cases, $\ApproachBoth$ reaches an F-measure of over 90\% for all entity types showing that the approach is generic and able to achieve good match quality for multiple heterogeneous entity types. Still there are some differences between the entity types. 
TVShows and Films generally perform worse than TVEpisodes and Persons with especially the precision for Film standing out negatively.
This is especially pronounced in the IMDB-TMDB and IMDB-TVDB datasets. This might be attributed to different sets of attributes between those datasets, e.g. as IMDB does not contain full-length descriptions of films and tv shows whereas TMDB and TVDB do.
Interestingly, Films/TVShows with very dissimilar titles due to different representations of non-English titles can be matched using the KGEs. For example the soviet drama "Defence Counsel Sedov" has the romanized title "Zashchitnik Sedov" in IMDB, while TMDB has either the translated "Defence Counsel Sedov" or the cyrillic \foreignlanguage{russian}{"Защитник Седов"}. These entity pairs are correctly matched in the $\ApproachBoth$ variant.  

To properly compare the performance of the approaches across all approaches we used the statistical analysis presented by Dem\v{s}ar~\cite{Demsar2006} and more specifically the Python package Autorank~\cite{autorank}, which aims to simplify the use of the proposed methods by Dem\v{s}ar. The performance measurement for each dataset and classifier are our paired samples. Given that we have more than two datasets, simply using hypothesis tests for all pairs would result in a multiple testing problem, which means the probability of accidentally reporting a significant difference would be highly increased. We therefore use the procedure recommended by Dem\v{s}ar: First we test if the average ranks of algorithms are significantly different using the Friedman test. If this is the case we perform a Nemenyi test to compare all classifiers and input combinations. 

\begin{figure}
    \centering
    \includegraphics[width=\columnwidth]{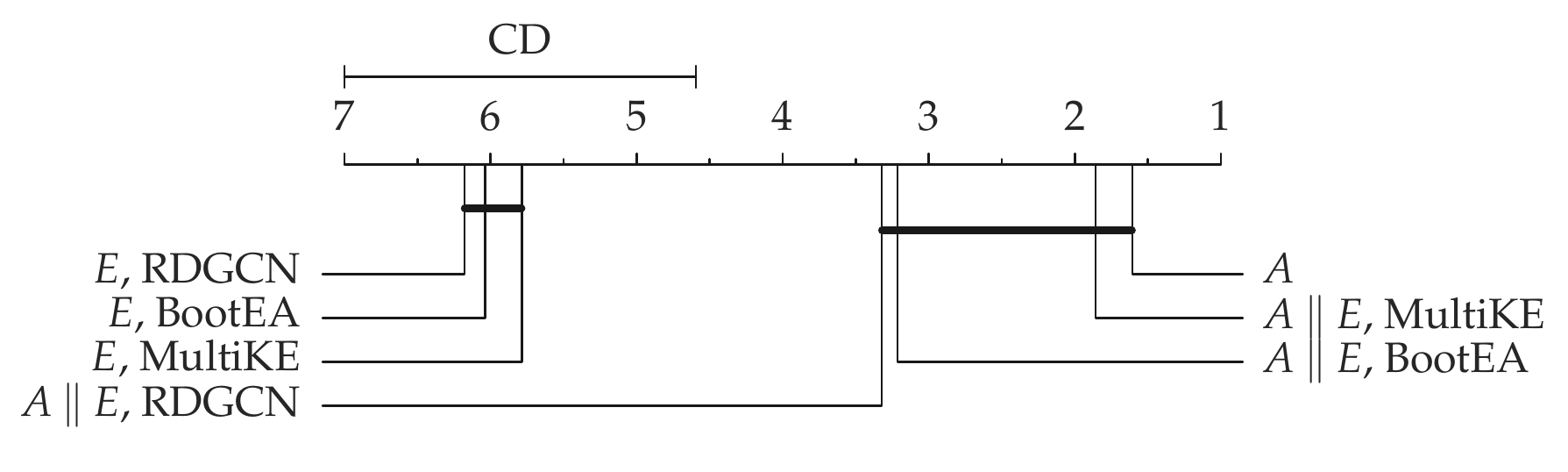}
    \caption{Critical distance diagram of Nemenyi test for shallow graph datasets, connected groups are not significantly different (at $p=0.05$)}
    \label{fig:cd_shallow}
\end{figure}

The null hypothesis of the Friedman test can be rejected ($p=1.60 \times 10^{-13}$). A Nemenyi test is therefore performed and we present the critical distance diagram in Figure~\ref{fig:cd_shallow}.
The axis shows the average rank of the input/embedding combination. Groups that are connected are not significantly different at the significance level of 0.05, which is internally corrected to ensure that all results together fulfill this.
Approaches that have a higher difference in average rank than the critical distance (CD) are significantly different.

While $\ApproachAttr$ performs the best, there is no significant difference to $\ApproachBoth$. What we can see is that $\ApproachEmb$ is significantly outperformed by all other approaches.

\subsection{Rich Graph Datasets}

\begin{table*}[]
    \caption{Averaged F-measure on test set of rich graph datasets. The best value in a row is highlighted. For average rank the best 3 values of the compared ranks are highlighted}
    \label{tab:richtable}
    \centering
\begin{tabular}{|c|c|cc|cc|cc|cc|cc|cc|cc|}
\hline
         \multicolumn{2}{|c|}{\multirow{3}{*}{Dataset}} & \multicolumn{6}{c|}{$\ApproachBoth$} & \multicolumn{2}{c|}{\multirow{2}{*}{$\ApproachAttr$}} & \multicolumn{6}{c|}{$\ApproachEmb$} \\
         \cline{3-8}
         \cline{11-16}
         \multicolumn{2}{|c|}{} & \multicolumn{2}{c|}{BootEA} & \multicolumn{2}{c|}{MultiKE} & \multicolumn{2}{c|}{RDGCN} & \multicolumn{2}{c|}{}& \multicolumn{2}{c|}{BootEA} & \multicolumn{2}{c|}{MultiKE} & \multicolumn{2}{c|}{RDGCN} \\
         \cline{3-16}
         \multicolumn{2}{|c|}{} &                        MLP &      RF &                                                                                MLP &     RF &    MLP &                         RF &           MLP &     RF &     MLP &      RF &     MLP &     RF &     MLP &      RF \\
\hline
\multirow{8}{*}{\rotatebox[origin=c]{90}{15K}} & D-W(V1) &                      0.775 &   0.668 &                                                          {\cellcolor{rankcolor1}}0.881 &  0.858 &  0.805 &                      0.842 &         0.827 &  0.828 &   0.764 &   0.678 &   0.853 &  0.871 &   0.718 &   0.707 \\
         & D-W(V2) &                      0.934 &   0.841 &                                                          {\cellcolor{rankcolor1}}0.945 &  0.918 &  0.897 &                      0.890 &         0.868 &  0.870 &   0.938 &   0.847 &   0.939 &  0.942 &   0.808 &   0.796 \\
         & D-Y(V1) &                      0.870 &   0.775 &                                                          {\cellcolor{rankcolor1}}0.986 &  0.982 &  0.974 &  {\cellcolor{rankcolor1}}0.986 &         0.972 &  0.971 &   0.837 &   0.746 &   0.952 &  0.941 &   0.947 &   0.953 \\
         & D-Y(V2) &                      0.983 &   0.908 &                                                          {\cellcolor{rankcolor1}}0.995 &  0.993 &  0.977 &                      0.991 &         0.978 &  0.978 &   0.975 &   0.888 &   0.973 &  0.971 &   0.947 &   0.960 \\
         & EN-DE(V1) &                      0.923 &   0.852 &                                                          {\cellcolor{rankcolor1}}0.986 &  0.984 &  0.966 &                      0.976 &         0.947 &  0.945 &   0.891 &   0.798 &   0.957 &  0.950 &   0.937 &   0.955 \\
         & EN-DE(V2) &                      0.970 &   0.918 &                                                          {\cellcolor{rankcolor1}}0.992 &  0.990 &  0.968 &                      0.978 &         0.956 &  0.955 &   0.946 &   0.875 &   0.961 &  0.958 &   0.934 &   0.956 \\
         & EN-FR(V1) &                      0.868 &   0.736 &                                                          {\cellcolor{rankcolor1}}0.978 &  0.973 &  0.950 &                      0.963 &         0.922 &  0.920 &   0.806 &   0.709 &   0.952 &  0.942 &   0.907 &   0.935 \\
         & EN-FR(V2) &                      0.965 &   0.876 &                                                          {\cellcolor{rankcolor1}}0.991 &  0.989 &  0.963 &                      0.977 &         0.937 &  0.936 &   0.942 &   0.875 &   0.977 &  0.978 &   0.921 &   0.948 \\
\cline{1-16}
\multirow{8}{*}{\rotatebox[origin=c]{90}{100K}} & D-W(V1) &                      0.873 &   0.850 &                                                          {\cellcolor{rankcolor1}}0.887 &  0.862 &  0.768 &                      0.774 &         0.810 &  0.811 &   0.868 &   0.820 &   0.850 &  0.871 &   0.645 &   0.556 \\
         & D-W(V2) &  {\cellcolor{rankcolor1}}0.962 &   0.927 &                                                                              0.951 &  0.923 &  0.756 &                      0.792 &         0.845 &  0.844 &   0.959 &   0.916 &   0.917 &  0.957 &   0.610 &   0.609 \\
         & D-Y(V1) &                      0.980 &   0.958 &                                                                              0.990 &  0.987 &  0.991 &  {\cellcolor{rankcolor1}}0.993 &         0.975 &  0.975 &   0.959 &   0.942 &   0.949 &  0.954 &   0.963 &   0.968 \\
         & D-Y(V2) &                      0.993 &   0.965 &                                                          {\cellcolor{rankcolor1}}0.995 &  0.990 &  0.983 &                      0.989 &         0.976 &  0.975 &   0.979 &   0.958 &   0.953 &  0.978 &   0.921 &   0.968 \\
         & EN-DE(V1) &                      0.943 &   0.907 &                                                          {\cellcolor{rankcolor1}}0.989 &  0.982 &  0.954 &                      0.961 &         0.944 &  0.943 &   0.901 &   0.859 &   0.956 &  0.947 &   0.872 &   0.891 \\
         & EN-DE(V2) &                      0.965 &   0.933 &                                                          {\cellcolor{rankcolor1}}0.993 &  0.988 &  0.926 &                      0.932 &         0.943 &  0.941 &   0.934 &   0.890 &   0.970 &  0.969 &   0.779 &   0.847 \\
         & EN-FR(V1) &                      0.925 &   0.867 &                                                          {\cellcolor{rankcolor1}}0.981 &  0.969 &  0.947 &                      0.938 &         0.920 &  0.919 &   0.866 &   0.819 &   0.948 &  0.943 &   0.866 &   0.894 \\
         & EN-FR(V2) &                      0.968 &   0.899 &                                                          {\cellcolor{rankcolor1}}0.989 &  0.979 &  0.897 &                      0.901 &         0.925 &  0.923 &   0.925 &   0.877 &   0.959 &  0.968 &   0.742 &   0.806 \\
\cline{1-16}
\multicolumn{2}{|c|}{Avg Rank} &                      5.938 &  11.094 &  {\cellcolor{avgrankcolor1}}\textcolor{white}{1.344} &  3.000 &  6.812 &                {\cellcolor{avgrankcolor2}}      5.375 &         7.688 &  8.281 &   8.625 &  12.625 &   6.125 & {\cellcolor{avgrankcolor3}} 5.656 &  11.969 &  10.469 \\
\hline
\end{tabular}
\end{table*}
The experiment results for the rich graph datasets are shown in Table~\ref{tab:richtable}.
It is evident that $\ApproachBoth$ achieves the best results. 
Overall it can solve the diverse match tasks including for multi-lingual KGs and  larger KGs very well with F-Measure values between 96\% and 99\% in most cases.
As before MultiKE performs the best out of all graph embedding approaches, especially in conjunction with the MLP classifier.

Comparing the performances between the datasets we see that on the variants with richer graph structure (V2) the results are better than on (V1) for the respective datasets. There is also a difference when contrasting the different sizes of the datasets. While $\ApproachBoth$ with BootEA and MultiKE generally seem to achieve better results on the larger 100K datasets compared to their 15K counterparts, this is less true for RDGCN.

\begin{figure}[]
    \centering
        \includegraphics[width=\columnwidth]{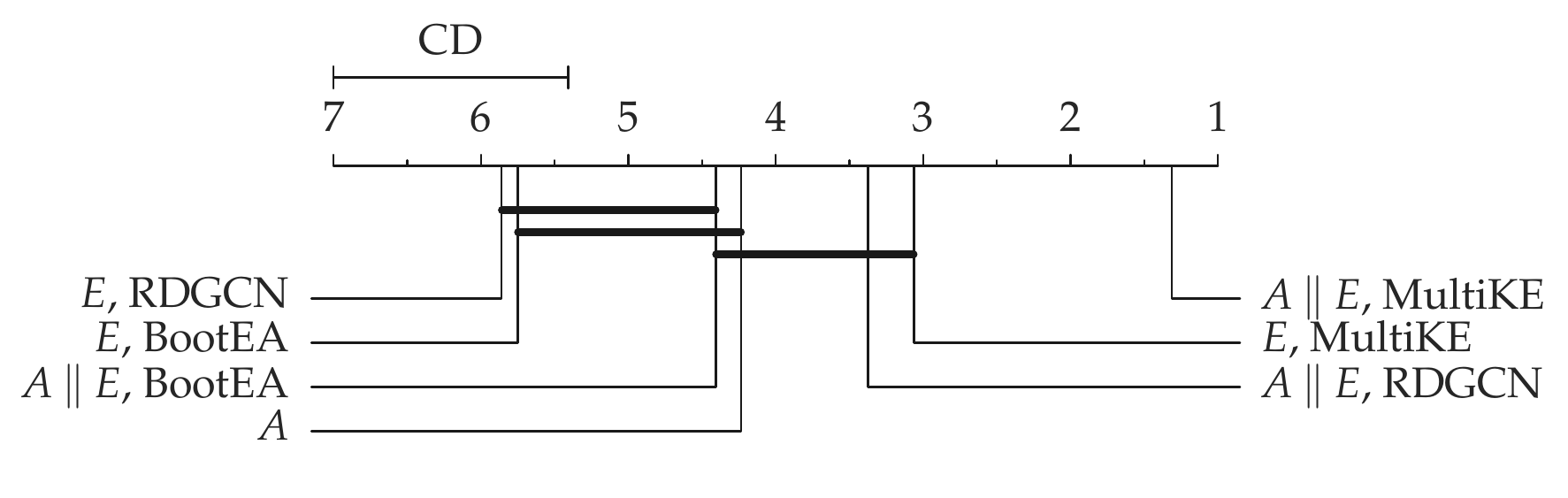}
    \caption{Critical distance diagram of Nemenyi test for rich graph datasets, connected groups are not significantly different (at $p=0.05$)}
    \label{fig:cd_rich}
\end{figure}

Again, we use the statistical procedure to make robust statements about performance. For our rich graph datasets we can reject the null hypothesis ($p=2.80 \times 10^{-20}$) of the Friedman test that all approaches and their average ranks should be equal. We therefore proceed and perform a Nemenyi test to determine which variants performed significantly different. The results are shown in Figure~\ref{fig:cd_rich}.
Generally, using $\ApproachBoth$ with any embedding approach performs better than $\ApproachEmb$, however for BootEA this difference is not significant. We can see that $\ApproachBoth$ with MultiKE is significantly better than all other variants. 
This is evidence that the combination of attribute similarities and embeddings is preferable to using attribute similarities or embeddings on their own for the task of entity resolution in rich knowledge graphs. This is even true for embedding techniques that already rely on attribute information such as MultiKE.
\subsection{Training Time}
\begin{table*}[]
\caption{Averaged training times (in seconds) on rich graph datasets of size 100K.}
 \label{tab:rich100ktimes}
 \centering
\begin{tabular}{|c|rr|rr|rr|}
\hline
 \multirow{2}{*}{Dataset}& \multicolumn{2}{c|}{$\ApproachEmb$} & \multicolumn{2}{c|}{$\ApproachAttr$} & \multicolumn{2}{c|}{$\ApproachBoth$} \\
 
\cline{2-7}
 & MLP & RF & MLP & RF & MLP & RF \\
\hline
D-W(V1) & 554.30 & 967.14 & 4,165.16 & 3,428.90 & 3,948.18 & 4,082.59 \\
D-W(V2) & 531.79 & 942.87 & 3,083.77 & 2,603.19 & 3,130.56 & 3,136.22 \\
D-Y(V1) & 380.90 & 809.90 & 938.38 & 242.76 & 570.24 & 699.73 \\
D-Y(V2) & 335.37 & 822.18 & 954.18 & 233.36 & 503.07 & 658.02 \\
EN-DE(V1) & 451.90 & 900.58 & 1,420.74 & 1,053.92 & 1,668.12 & 1,688.45 \\
EN-DE(V2) & 334.95 & 898.91 & 1,064.44 & 775.35 & 1,279.61 & 1,365.34 \\
EN-FR(V1) & 456.43 & 858.63 & 2,183.16 & 1,755.25 & 2,263.96 & 2,360.49 \\
EN-FR(V2) & 377.26 & 819.43 & 1,642.92 & 1,281.80 & 1,870.10 & 1,806.68 \\
\hline
\end{tabular}
\end{table*}

Experiments were run on a cluster provided by the Leipzig University Computing Center, which is comprised of several nodes with AMD EPYC 32 core processors and up to 512GB RAM. Experiments were run on a single node. To illustrate the relative runtimes of the considered variants we focus on the bigger KGs with 100K entities. 
Table \ref{tab:rich100ktimes} shows the running times for training on each 100K dataset, averaged over all 5 folds. The full training times are mostly dominated by the pre-processing of the attribute similarities. This pre-processing is not necessary for $\ApproachEmb$ and hence training time is up to about 8 times faster for MLP.
On average, training times for $\ApproachBoth$ are slightly longer than for $\ApproachAttr$ due to an increase in the dimensionality of the input.

\subsection{Comparison with other approaches}
We compare our approach to the state-of-the-art ER frameworks Magellan~\cite{Magellan} and DeepMatcher~\cite{DeepMatcher}. Magellan is an ER framework that allows the  use of ML classifiers for ER. We present the best performing classifiers XGBoost~\cite{XGBoost} and Random Forest (rf). DeepMatcher provides several deep learning solutions for ER, we employ the \textit{hybrid} variant which uses a bidirectional recurrent neural network with a decomposable attention-based attribute summarization module.
To avoid any decrease in performance due to blocking we provide both frameworks with respective training or test entity mappings directly. 
Because such a setup is not possible for the approaches discussed in~\cite{Sun2020}, which mostly use resolution strategies based on nearest neighbors, we cannot fairly compare our approach with theirs and therefore refrain from this comparison here.

\subsubsection{Shallow graph datasets}
\begin{figure}[]
    \centering
    \includegraphics[width=\columnwidth]{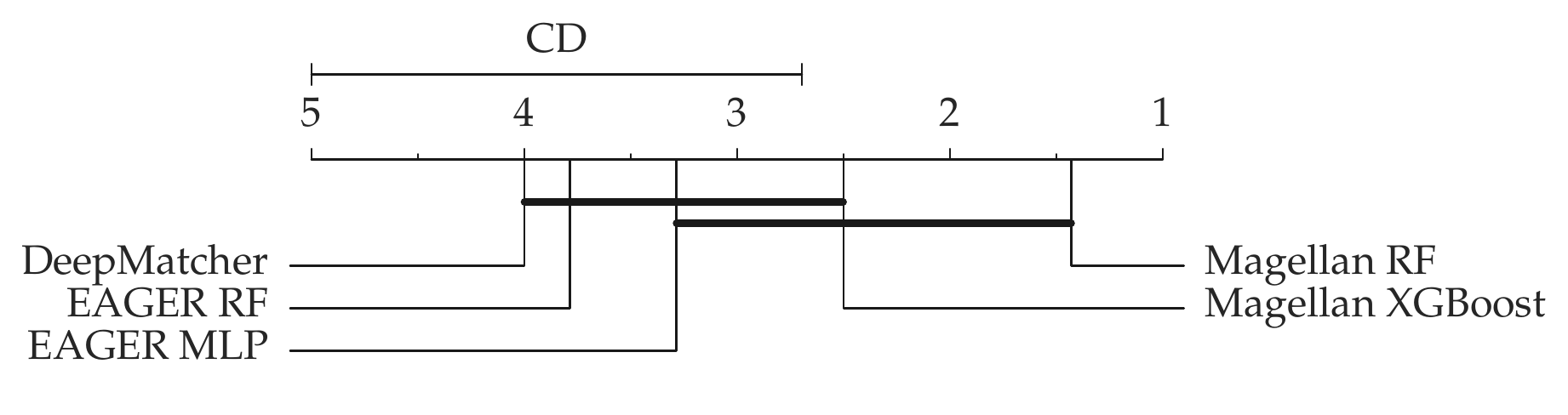}
    \caption{Critical distance diagram of Nemenyi test for shallow graph datasets, connected groups are not significantly different (at $p=0.05$)}
    \label{fig:cd_magellan}
\end{figure}

We start with the comparison for the shallow datasets.
Since both Magellan and DeepMatcher expect matched schemata we align the attributes by hand where necessary. We report F-measure (fm), Precison (prec) and Recall (rec) averaged over the 5 folds, with the variance over the folds in Table~\ref{tab:magellantable}.
For the comparison with other approaches we use $\ApproachBoth$ and for brevity we will refer to it simply as EAGER.
\begin{table*}[]
    \caption{Averaged F-measure, precision and recall on test set of shallow graph datasets. The best F-measure in a row is highlighted}
    \label{tab:magellantable}
    \centering
    
\resizebox{\textwidth}{!}{%
\begin{tabular}{|c|ccc|ccc|ccc|ccc|ccc|}
\hline
\multirow{2}{*}{Dataset} & \multicolumn{3}{c|}{EAGER MLP} & \multicolumn{3}{c|}{EAGER RF} & \multicolumn{3}{c|}{DeepMatcher} & \multicolumn{3}{c|}{Magellan XGBoost} & \multicolumn{3}{c|}{Magellan RF} \\
\cline{2-16}
      &               fm &   prec &    rec &       fm &   prec &    rec &          fm &   prec &    rec &               fm &   prec &    rec &                         fm &   prec &    rec \\
\hline
abt-buy       &                      0.958 &  0.975 &  0.942 &    0.952 &  0.963 &  0.941 &       0.930 &  0.885 &  0.980 &            0.974 &  0.971 &  0.978 &  {\cellcolor{rankcolor1}}0.977 &  0.975 &  0.979 \\
amazon-google &  {\cellcolor{rankcolor1}}0.789 &  0.794 &  0.787 &    0.760 &  0.804 &  0.722 &       0.743 &  0.673 &  0.836 &            0.724 &  0.737 &  0.712 &                      0.727 &  0.766 &  0.693 \\
dblp-acm      &                      0.997 &  0.999 &  0.994 &    0.997 &  0.999 &  0.995 &       0.990 &  0.980 &  0.999 &            0.998 &  0.999 &  0.997 &  {\cellcolor{rankcolor1}}0.999 &  1.000 &  0.998 \\
dblp-scholar  &                      0.994 &  0.997 &  0.990 &    0.997 &  0.999 &  0.996 &       0.994 &  0.992 &  0.997 &            0.997 &  0.997 &  0.997 &  {\cellcolor{rankcolor1}}0.998 &  0.998 &  0.997 \\
imdb-tmdb     &                      0.988 &  0.985 &  0.992 &    0.984 &  0.978 &  0.990 &       0.984 &  0.971 &  0.997 &            0.995 &  0.998 &  0.993 &  {\cellcolor{rankcolor1}}0.997 &  0.997 &  0.996 \\
imdb-tvdb     &                      0.973 &  0.959 &  0.987 &    0.967 &  0.940 &  0.994 &       0.987 &  0.979 &  0.996 &            0.993 &  0.992 &  0.993 &  {\cellcolor{rankcolor1}}0.994 &  0.991 &  0.996 \\
tmdb-tvdb     &                      0.983 &  0.989 &  0.977 &    0.981 &  0.991 &  0.971 &       0.988 &  0.978 &  0.998 &            0.993 &  0.992 &  0.994 &  {\cellcolor{rankcolor1}}0.995 &  0.993 &  0.997 \\
\hline
Avg Rank & \multicolumn{3}{c|}{{\cellcolor{avgrankcolor3}}3.286} & \multicolumn{3}{c|}{3.786} & \multicolumn{3}{c|}{4.000} & \multicolumn{3}{c|}{{\cellcolor{avgrankcolor2}}2.500} & \multicolumn{3}{c|}{{\cellcolor{avgrankcolor1}}\textcolor{white}{1.429}}\\ 
\hline
\end{tabular}
}
\end{table*}
All frameworks perform very well with almost all F-measure values over $0.95$ except on amazon-google.
Magellan achieves higher f-measures than EAGER on all datasets except amazon-google.
The statistical analysis shows a significant difference: $p=0.012$ using the Friedman test. However, looking at the critical distance diagram in Figure~\ref{fig:cd_magellan} we can see that only DeepMatcher and EAGER RF is significantly outperformed by Magellan RF. There is no significant difference between EAGER MLP and Magellan RF but EAGER does not depend on the provision of schema matching.
\subsubsection{Rich graph datasets}
\begin{figure}[]
    \centering
    \includegraphics[width=\columnwidth]{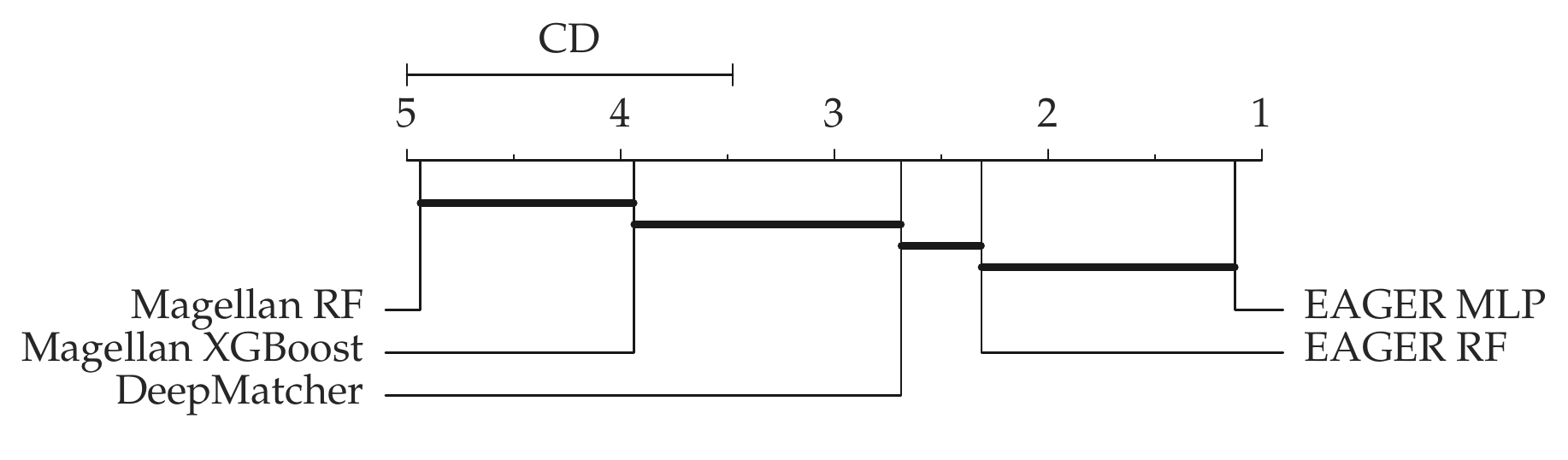}
    \caption{Critical distance diagram of Nemenyi test for rich graph datasets, connected groups are not significantly different (at $p=0.05$)}
    \label{fig:magellan_rich_cd}
\end{figure}

For the rich graph datasets the heterogeneity of the different KGs was a problem for Magellan and DeepMatcher since they both expect perfectly matched schemata. This was manageable for the smaller datasets, where this can be done by hand. In order to use Magellan and Deepmatcher on the rich graph datasets we did the same as for EAGER and concatenated all entity attributes into a single attribute.
\begin{table*}[]
    \caption{Averaged F-measure, Precision and Recall on test set of rich graph datasets. The best F-measure value in a row is highlighted}
    \label{tab:richftablemagellan}
    \centering
\resizebox{\textwidth}{!}{%
\begin{tabular}{|c|c|ccc|ccc|ccc|ccc|ccc|}
\hline
    \multicolumn{2}{|c|}{\multirow{2}{*}{Dataset}} & \multicolumn{3}{c|}{EAGER MLP} & \multicolumn{3}{c|}{EAGER RF} & \multicolumn{3}{c|}{DeepMatcher} & \multicolumn{3}{c|}{Magellan XGBoost} & \multicolumn{3}{c|}{Magellan RF} \\
    \cline{3-17}
    \multicolumn{2}{|c|}{}  &                         fm &   prec &    rec &                         fm &   prec &    rec &                         fm &   prec &    rec &               fm &   prec &    rec &          fm &   prec &    rec \\
\hline
\multirow{8}{*}{\rotatebox[origin=c]{90}{15K}} & D-W(V1) &  {\cellcolor{rankcolor1}}0.881 &  0.990 &  0.794 &    0.858 &  0.991 &  0.756 &                      0.876 &  0.854 &  0.899 &            0.837 &  0.896 &  0.786 &       0.822 &  0.849 &  0.798 \\
    & D-W(V2) &  {\cellcolor{rankcolor1}}0.945 &  0.993 &  0.903 &    0.918 &  0.992 &  0.854 &                      0.904 &  0.895 &  0.914 &            0.863 &  0.913 &  0.818 &       0.848 &  0.867 &  0.830 \\
    & D-Y(V1) &  {\cellcolor{rankcolor1}}0.986 &  1.000 &  0.972 &    0.982 &  1.000 &  0.964 &                      0.980 &  0.976 &  0.984 &            0.973 &  0.975 &  0.970 &       0.972 &  0.973 &  0.971 \\
    & D-Y(V2) &  {\cellcolor{rankcolor1}}0.995 &  1.000 &  0.991 &    0.993 &  0.999 &  0.987 &                      0.987 &  0.984 &  0.990 &            0.975 &  0.977 &  0.972 &       0.974 &  0.975 &  0.974 \\
    & EN-DE(V1) &  {\cellcolor{rankcolor1}}0.986 &  0.997 &  0.974 &    0.984 &  0.996 &  0.971 &                      0.968 &  0.972 &  0.964 &            0.966 &  0.990 &  0.944 &       0.960 &  0.977 &  0.945 \\
    & EN-DE(V2) &  {\cellcolor{rankcolor1}}0.992 &  0.997 &  0.988 &    0.990 &  0.997 &  0.982 &                      0.975 &  0.973 &  0.977 &            0.973 &  0.992 &  0.955 &       0.970 &  0.985 &  0.955 \\
    & EN-FR(V1) &  {\cellcolor{rankcolor1}}0.978 &  0.996 &  0.960 &    0.973 &  0.994 &  0.952 &                      0.954 &  0.950 &  0.959 &            0.953 &  0.984 &  0.924 &       0.951 &  0.979 &  0.924 \\
    & EN-FR(V2) &  {\cellcolor{rankcolor1}}0.991 &  0.997 &  0.984 &    0.989 &  0.996 &  0.982 &                      0.968 &  0.965 &  0.972 &            0.971 &  0.993 &  0.949 &       0.970 &  0.993 &  0.949 \\
\cline{1-17}
\multirow{8}{*}{\rotatebox[origin=c]{90}{100K}} & D-W(V1) &                      0.887 &  0.994 &  0.801 &    0.862 &  0.989 &  0.764 &  {\cellcolor{rankcolor1}}0.925 &  0.905 &  0.946 &            0.817 &  0.904 &  0.746 &       0.815 &  0.887 &  0.754 \\
    & D-W(V2) &  {\cellcolor{rankcolor1}}0.951 &  0.991 &  0.915 &    0.923 &  0.988 &  0.866 &                      0.929 &  0.912 &  0.947 &            0.834 &  0.922 &  0.761 &       0.830 &  0.892 &  0.775 \\
    & D-Y(V1) &                      0.990 &  1.000 &  0.981 &    0.987 &  1.000 &  0.976 &  {\cellcolor{rankcolor1}}0.992 &  0.991 &  0.993 &            0.983 &  0.991 &  0.976 &       0.982 &  0.986 &  0.979 \\
    & D-Y(V2) &  {\cellcolor{rankcolor1}}0.995 &  1.000 &  0.991 &    0.990 &  1.000 &  0.982 &                      0.993 &  0.992 &  0.995 &            0.985 &  0.987 &  0.983 &       0.984 &  0.984 &  0.983 \\
    & EN-DE(V1) &  {\cellcolor{rankcolor1}}0.989 &  0.997 &  0.981 &    0.982 &  0.997 &  0.968 &                      0.972 &  0.974 &  0.971 &            0.967 &  0.991 &  0.945 &       0.966 &  0.987 &  0.946 \\
    & EN-DE(V2) &  {\cellcolor{rankcolor1}}0.993 &  0.997 &  0.990 &    0.988 &  0.997 &  0.980 &                      0.977 &  0.975 &  0.980 &            0.969 &  0.993 &  0.945 &       0.966 &  0.985 &  0.947 \\
    & EN-FR(V1) &  {\cellcolor{rankcolor1}}0.981 &  0.996 &  0.966 &    0.969 &  0.995 &  0.946 &                      0.956 &  0.958 &  0.955 &            0.947 &  0.988 &  0.908 &       0.945 &  0.984 &  0.910 \\
    & EN-FR(V2) &  {\cellcolor{rankcolor1}}0.989 &  0.994 &  0.983 &    0.979 &  0.992 &  0.967 &                      0.968 &  0.966 &  0.970 &            0.963 &  0.991 &  0.937 &       0.961 &  0.987 &  0.937 \\
\hline
\multicolumn{2}{|c|}{Avg Rank}  & \multicolumn{3}{c|}{{\cellcolor{avgrankcolor1}}\textcolor{white}{1.125}} & \multicolumn{3}{c|}{{\cellcolor{avgrankcolor2}}2.312} & \multicolumn{3}{c|}{{\cellcolor{avgrankcolor3}}2.688} & \multicolumn{3}{c|}{3.938} & \multicolumn{3}{c|}{4.938}\\ 
\hline
\end{tabular}
}
\end{table*}
We can see in Table~\ref{tab:richftablemagellan} that EAGER using MLP outperforms all other approaches except on D-W (V1) and D-Y (V1) for the 100K sizes, where DeepMatcher performs best. Magellan is outperformed on all datasets by EAGER and DeepMatcher. Contrary to the smaller datasets the bigger number of training examples seems especially beneficial for DeepMatcher. Using our statistical analysis we can reject the Friedman test ($p=2.21 \times 10^{-11}$) and therefore show the results of the Nemenyi tests in Figure~\ref{fig:magellan_rich_cd}.
It is apparent, that our approach significantly outperforms Magellan, while EAGER using MLP also significantly outperforms DeepMatcher overall.
\section{Conclusion \& Future work}\label{sec:conclusion}
We explored the combination of knowledge graph embeddings and attribute similarities for entity resolution in knowledge graphs. These approaches are included in a new learning-based ER system called EAGER. We tested our approach on a range of different datasets and showed that using a combination of both graph embeddings and attribute similarities generally yields the best results compared to just using either one. We showed that our approach yields competitive results that are on par with or significantly outperform state of the art approaches. The approach is generic and can deal with arbitrary entity types without prior schema matching. 

In the future we will investigate blocking strategies based on both embeddings and attribute information to improve runtimes and thus scalability to large  datasets. 
We will also explore whether new property matching schemes like LEAPME \cite{ayala2020leapme} can be utilized for blocking to reduce the high cost in pre-processing attribute similarities.
As training data in practice is often too small or hard to obtain at all, using alternative learning strategies such as unsupervised and active learning in this context should be explored.
\bibliographystyle{IEEEtran}
\bibliography{library}
\end{document}